# End-to-End Change Detection for High Resolution Drone Images with GAN Architecture

Yura Zharkovsky, Ovadya Menadeva Percepto
`{yura,ovadya}@percepto.co`

## Abstract

Abstract: Monitoring large areas is presently feasible with high resolution drone cameras, as opposed to time-consuming and expensive ground surveys. In this work we reveal for the first time, the potential of using a state-of-the-art change detection GAN based algorithm with high resolution drone images for infrastructure inspection. We demonstrate this concept on solar panel installation. A deep learning, data-driven algorithm for identifying changes based on a change detection deep learning algorithm was proposed.
We use the Conditional Adversarial Network approach to present a framework for change detection in images. The proposed network architecture is based on pix2pix GAN framework.   Extensive experimental results have shown that our proposed approach outperforms the other state-of-the-art change detection methods.

**1. Introduction**

Change detection in time-varying sequences of drone or satellite images acquired in the same geographical area is an important part of many practical applications, including solar panel installation monitoring or urban development analysis, environmental inspection and agricultural monitoring. In most cases, solving the change detection task in manual mode is a highly time-consuming operation. At present, the best results in the overwhelming majority of image analysis and processing tasks are delivered by methods based on deep convolutional neural networks (CNN). In this paper, we propose an improved method for automatic change detection in drone and remote sensing images, which employs an advanced type of CNN Conditional Adversarial Networks. We improved on previous work by applying ResNet 9 architecture to the generator.

**2. Related Works**

Remote sensing application object-based pixel wise detection as described in [6]. The pixel-wise change detection contains the direct, transform based framework, which is based on post-classification [2] and presumes extracting objects and independently classifying them [5]. The method described here does not use object classification. It calculates the change detection directly on the image using GAN similar to [1].

## 3. Methodology

In this work, we consider change detection only in image differences that correspond to real changes in a scene - for example due to construction or other human activities such as solar panel installations or placement of new objects - rather than differences due to season-specific changes, brightness variations and other factors (see Figure 1).

We propose using a Generative Adversarial Network GAN similar to [1]. We compare the feature space using domain adaptation with pix2pix as described in [7].

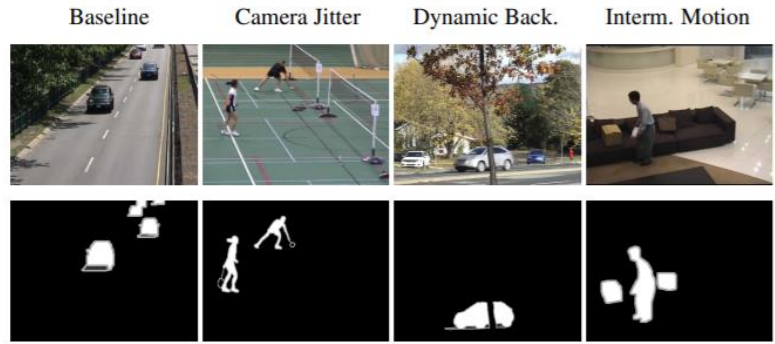

**Figure 1.** CDnet 2014: An Expanded Change Detection Benchmark as described in [13]. Changes due to object movement such as cars and seasonal changes of vegetation are learned by the algorithm as non-changes.

### 3.1 Problem Statement

In a basic generative model (GAN), there is no control over instances of the data being generated. In our case, we use Conditional GAN (CGAN), see [1], and the generator learns to generate a fake sample with a specific condition such as a label associated with an image or more detail, rather than a generic sample from unknown noise distribution. The generator G, based on the features object space from input data x synthesizes output data y.

G : {x, z} → y

The discriminator D learns how to detect 'fake' images generated by the generator G

D : {x,y} → p

The discriminator maps objects from the data space to [0,1] p interval, which is interpreted as the probability that the example was "real". As a result, D and G play the following two-player minimax game:

$L_{cGAN}(G, D) = E_{x,y}[\log D(x, y)] + E_{x,z}[\log(1 - D(x, G(x, z)))]$

## 3.2 GAN architectures

In the implementation, the generator applies the transformations to a pair of input images simultaneously and extracts features from these images similar to Siam networks. In order to do this, the concatenation procedure is applied to the input images of the generator. The generator is based on the ResNet 9 network (see figure 3 [13]). The discriminator architecture is based on "PatchGAN" [7]. The input to the discriminators are three images: two images for comparison and the third image the label of changes map - this is either the generator output or ground truth change detection labels (see figure 2). The discriminator is trained to separate the change detection map generated by the generator and ground truth labels.

In order to train the GAN, which is basically the generator and the discriminator, they are trained together. The generator generates the change detection map, the discriminator decides whether this change detection map is fake or real for two input images similar to [1]. The training pipeline of the discriminator is shown on Figure 2.

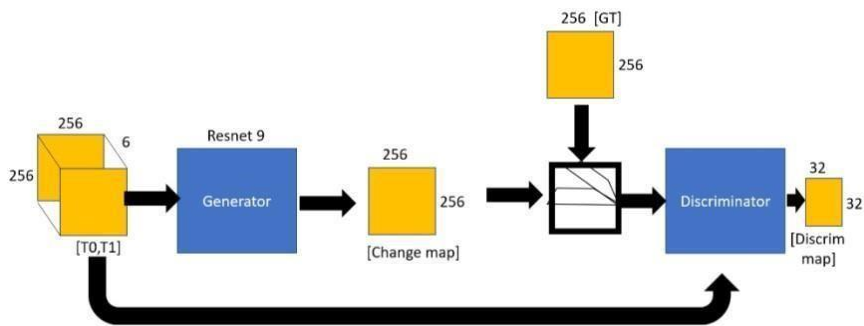

**Figure 2.** The training pipeline of the Conditional Adversarial Network

At the next training step, the generator parameters are updated using classification error that uses the discriminator output and the discrepancy between the difference map and the ground truth labels. The training pipeline of the generator is shown on Figure 3.

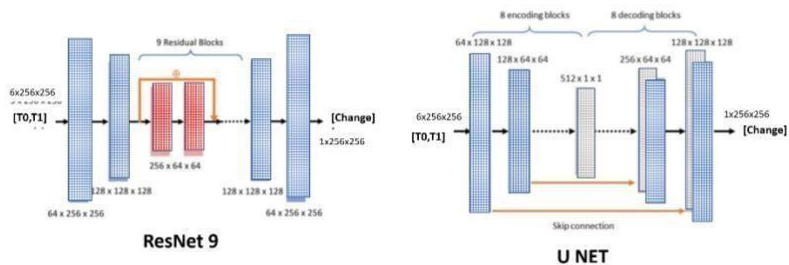

**Figure 3.** The generator backbone ResNet 9 versus U-NET

## 4. Experiments

In this work, we use transfer learning with simple examples of simulated data in order to test the proposed network. We simulate change detection using synthetic images that use simple polygons with small relative shift of objects, and change detection in real images. And finally, solar panel installation monitoring using high resolution images from Precepto Drones demonstrates the detection of actual construction changes.

### 4.1 Experiments on the Simulated Data

The AICD Dataset Change Detection dataset contains 1000 pairs of 800x600 images, each pair consisting of one reference image and one test image, and the 1000 corresponding 800x600 ground truth masks. The images were rendered using the realistic rendering engine of the game Virtual Battle Station 2. The dataset consists of 100 different scenes containing several objects (trees, buildings, etc.) and moderate ground relief. At approximately 250 meters high, the results of our algorithm running on the AICD Dataset:

On the validation set data statistics: Recall 85.1%, Precision 95.40%

Compared to the (Recall) 83% (Precision) 94% in the baseline by [1]

### 4.2 Experiments on the Real Image Dataset

In the second type of experiments, a neural-network architecture was evaluated using real images. For dataset we used solar panel construction, using drone mapping images of the same region 3 months apart (see figure 4). We obtained 50 pairs of season-varying images with resolution of 4096x2160 pixels for manual ground truth creation. That allowed us to take into account objects with different sizes (i.e. from boxes to big construction structures), season changes of natural objects (i.e. from single trees to wide forest areas). Dataset was generated by randomly cropping 256x256 patches. We used Intersection over Union (IoU) metrics to assess change detection quality. Additionally, we applied Precision and Recall metrics on the entire test dataset. For IoU thresholds equal to 0.5, the average Precision and Recall values were 0.92 and 0.93, respectively.

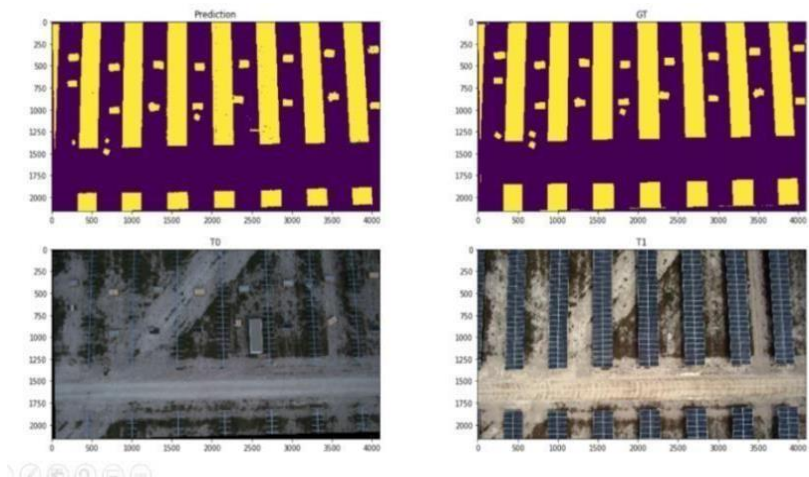

**Figure 4.** An example of solar panel installation monitoring using drone images: bottom left – input image A, bottom right – input image B, top left – synthesized difference map, top right – ground truth

### 5. Conclusion

This work demonstrates how to apply a conditional Generative Adversarial Network (GAN) "pix2pix" architecture for automatic change detection in drone and remote sensing images. An extensive database of synthetic and real images was used. The performed tests have shown that the proposed CNN is promising and efficient enough in change detection on synthetic and real images. We demonstrate this algorithm on images using high resolution drone data for solar panel installation monitoring.

### 6. Acknowledgement

We would like to Thank Inbal Gamliel for her help in gathering the data and providing the annotation. And to Hezi Roda , Ariel Benitah for providing the Alignment algorithm and fruitful discussions.